\documentclass[a4paper,conference]{IEEEtran}
\usepackage{cite}
\def\etal{\emph{et al}}
\ifCLASSINFOpdf
   \usepackage[pdftex]{graphicx}
   \graphicspath{{pdf/}{images/}}
   \DeclareGraphicsExtensions{.pdf,.jpeg,.jpg,.png}
\else
   \usepackage[dvips]{graphicx}
   \graphicspath{{../eps/}}
   \DeclareGraphicsExtensions{.eps}
\fi
\usepackage{amsmath}
\usepackage{amssymb}
\usepackage{array}
\ifCLASSOPTIONcompsoc
  \usepackage[caption=false,font=normalsize,labelfont=sf,textfont=sf]{subfig}
\else
  \usepackage[caption=false,font=footnotesize]{subfig}
\fi
\usepackage{url}
\usepackage{multirow}
\usepackage{adjustbox}

\begin{document}
\title{ReADS: A Rectified Attentional Double Supervised Network for Scene Text Recognition}

%\author{
%\IEEEauthorblockN{Qi Song}
%\IEEEauthorblockA{Meituan-Dianping Group\\
%Beijing, China\\
%Email: songqi03@meituan.com}
%\and
%\IEEEauthorblockN{Qianyi Jiang}
%\IEEEauthorblockA{Meituan-Dianping Group\\
%Beijing, China\\
%Email: jiangqianyi02@meituan.com}
%\and
%\IEEEauthorblockN{Nan Li}
%\IEEEauthorblockA{Meituan-Dianping Group\\
%Beijing, China\\
%Email: linan21@meituan.com}
%\and
%\IEEEauthorblockN{Rui Zhang}
%\IEEEauthorblockA{Meituan-Dianping Group\\
%Beijing, China\\
%Email: zhangrui36@meituan.com}
%\and
%\IEEEauthorblockN{Xiaolin Wei}
%\IEEEauthorblockA{Meituan-Dianping Group\\
%Beijing, China\\
%Email: weixiaolin02@meituan.com}
%}


\author{\IEEEauthorblockN{Qi Song,
Qianyi Jiang,
Nan Li,
Rui Zhang and
Xiaolin Wei}
\IEEEauthorblockA{Meituan-Dianping Group,\\
Beijing, China\\ 
Email: \{songqi03,jiangqianyi02,linan21,zhangrui36,weixiaolin02\}@meituan.com}
}


\maketitle

\begin{abstract}
In recent years, scene text recognition is always regarded as a sequence-to-sequence problem. Connectionist Temporal
Classification (CTC)  and Attentional sequence recognition (Attn) are two very prevailing approaches to tackle this problem while they may fail in some scenarios respectively. CTC concentrates more on every individual character but is weak in text semantic dependency modeling. Attn based methods have better context semantic modeling ability while tends to overfit on limited training data. In this paper, we elaborately design a Rectified Attentional Double Supervised Network (ReADS) for general scene text recognition. To overcome the weakness of CTC and Attn, both of them are applied in our method but with different modules in two supervised branches which can make a complementary to each other. Moreover, effective spatial and channel attention mechanisms are introduced to eliminate background noise and extract valid foreground information. Finally, a simple rectified network is implemented to rectify irregular text. The ReADS can be trained end-to-end and only word-level annotations are required. Extensive experiments on various benchmarks verify the effectiveness of ReADS which achieves state-of-the-art performance.
%Then the ReADS can success in many difficult situations which fails  before. 
\end{abstract}


% \fi
\IEEEpeerreviewmaketitle

\section{Introduction}
%1.介绍scene text背景，重要性
%2.最近大部分进展都是基于ctc or seq2seq，举例：xxx
% 但ctc缺乏上下文语义（即使加入了一个rnn，效果也不够好），seq2seq上下文建模过强，数据不够时容易过拟合
%3.在本文中，我们考虑把两者结合，构造一个双分支网络，
%同时，为了矫正弯曲文字，可以加入tps
%为了去除背景噪声干扰，提取有效前景信息，加入注意力机制
%最后，总的来说，我们的贡献包含如下几点：
%1.引入双分支，即引入上下文又关注纹理信息
%2.引入有效的注意力机制
%3.两个分支结果融合？
%4.验证在规则和不规则均有效，后续可以考虑引入二维注意力，更好的解决弯曲文字问题

Scene text recognition is an important computer vision task that reading text from images. It is an indispensable component for image understanding from high-level semantic information retrieving. Many challenging applications such as license reading, table analyzing and document processing, benefit from the maturity of Optical Character Recognition (OCR). However, scene text recognition remains an unsolved problem because of drastic variations in appearance, illumination, noise, layout, and background.

Recent advances in scene text recognition are mostly inspired by the success of deep learning techniques. Among them,Connectionist Temporal Classification (CTC) and Attentional sequence recognition (Attn) are the two most popular methods incorporated with Convolutional Neural Networks (CNNs), Recurrent Neural Networks (RNNs) or some other basic deep learning components, to form a framework named encoder-decoder. 
Both of them can handle text images with variable length. Attn based approaches are more accurate in most scenarios while CTC based ones, such as \cite{shi2017crnn,liu2016starnet,Gao2017ACSM}, achieve better efficiency and are easier to train.
However, some drawbacks still exist in these two mainstream methods respectively. 
Firstly, CTC based approaches predict all characters in one time so the sequential and semantic dependency between output characters is not modeled explicitly. The performance of CTC degrades when part of the text images are polluted by illumination, noise or some other reasons. Even combined with RNNs in the encoding state, the decoder still suffers from lacking semantic context between characters. Therefore, predefined lexicons are always needed to refine the output of CTC based approaches.
Secondly, Attn based methods can embed the language model into the attentional recurrent decoder to depict character dependencies. However, in Natural Language Processing (NLP), to construct a reasonable language model, massive data (often billions) is indispensable. Two largest scene text dataset\cite{corr2014Jaderberg,Gupta2016CVPR} only contain millions of text images, besides the images are synthesized. Not surprisingly, Attn based methods often tend to overfit on the limited training data and remain unsatisfied on the real world scene text recognition.

%Then in the encoder, the backbone is an attentional CNN which can distinguish disturbed backgrounds and highlight valid foregrounds shared by the CTC and Attn branches.
In this paper, to deal with the problems mentioned above, a \textbf{Re}cified \textbf{A}ttentional \textbf{D}ouble \textbf{S}upervised network named ReADS is proposed as shown in Figure \ref{framework}. Our method first rectifies the input images by adopting a Spatial Transformer Network (STN) as in ~\cite{NIPS2015stn}. Then in the encoder, the backbone is built up by attentional CNNs shared by the CTC and the Attn branches. Moreover, a multi-layer Bidirectional LSTM\cite{Hochreiter1997lstm} is also adopted for the attention branch. Finally, in the decoder, both CTC and Attn are applied as double supervisions. The CTC branch mainly concentrates on visual feature representation inference and the Attn branch relies on semantic context modeling of characters. Thus our method give the final recognition result from two different views in a decoupled style. 
%To make the most of them, the CTC branch only concentrates on text recognition by image feature and texture representation. On the contrary, in the Attn branch, RNNs are equipped in both encoding and decoding states to improve semantic context modeling.
% It is worth noting that, to make the most of them, there are no RNNs in the CTC branch to avoid introduce the sequential and semantic dependencies of characters. 

In summary, our main contributions are as follows:
\begin{itemize}
	\item [1)] 
	We propose a novel double supervised network which predicts text from both image inherent texture and semantic context by CTC and Attn. The proposed method can overcome the shortcomings in previous single supervised approaches and achieve better accuracy.
	\item [2)]
	A simple but effective attention mechanism is applied in the encoder, which discriminates foreground text features from messy backgrounds. A rectified module is also used in front of the encoder for handling irregular text images.
	\item [3)]
	Our proposed method achieves the state-of-the-art performance on both regular and irregular scene text benchmarks. Especially, our method is only trained on synthetic data and no real world text images are used.
\end{itemize}

\section{Related work}
%早期方法主要用于文档识别，用了一些滑窗、笔划检测等技术，随着深度学习技术的兴起，大部分方法都基于一种非对称encoder-decoder结构，其中encoder一般是一个基于cnn和rnn的神经网络，而decoder则主要由两种方法构成，即CTC和Attn
Early works in text recognition mainly focus on document text. Some handcrafted features, such as connected components~\cite{c_c} or Hough Voting~\cite{Yao_2014_CVPR}, are incorporated with the sliding window technique ~\cite{15549WordSpotting,e2e} to detect and recognize every single character. Then a graph-based inference is utilized to find words with the max probability from the characters. However, such kinds of traditional methods can not handle some intractable scenarios such as scene text because of noisy background, various appearance and uncontrollable illumination. With the rise of deep learning, recent approaches tend to treat text recognition as a sequence-to-sequence problem. And most of them are based on an asymmetrical encoder-decoder framework. Typically, the encoder consists of some CNNs and RNNs while the decoder mainly relies on two implementations, namely CTC and Attn.
\subsection{CTC based Text Recognition}
%1.ctc介绍,2.基于ctc的方法介绍
%readed_2015_An End-to-End Trainable Neural Network for Image-based Sequence Text Recognition
%readed_2016_Fully Convolutional Recurrent Network for Handwritten Chinese Text Recognition
%readed_2016_STAR-Net- A SpaTial Attention Residue Network for Scene Text Recognition
%readed_2018.08_Rosetta Large Scale System for Text Detection and Recognition in Images
CTC~\cite{graves2006ctc} is originally proposed by Graves \etal. for speech recognition. Since both speech and text recognition can be regarded as the sequence-to-sequence problems, CTC is prevailing in recent scene text recognition researches. In ~\cite{shi2017crnn}, an end-to-end trainable network named CRNN is proposed to directly recognize text lines without character level annotations. CRNN encodes text images using CNNs and bi-directional RNNs, then decodes by CTC. Xie \etal.~\cite{xie2016fcrn} utilize CTC to deal with online handwritten Chinese text recognition. The proposed method first transforms input trajectories into fixed size images, then the images are fed into a network whose structure is similar to CRNN. Liu \etal.~\cite{liu2016starnet} present the STAR-Net which goes one step further. To achieve a better performance, the STAR-Net employs a spatial transformer to rectify input text images and a deeper CNN using residual structures is adopted to enhance the model representation. In industrial applications, Facebook conducts an OCR system called \textit{Rosetta} ~\cite{Rosetta} to process daily uploaded images. To compromise between effectiveness and efficiency, the bi-directional RNNs before CTC are removed and a complicated training strategy is invented to fit the massive real world data. It should be noted that the approaches mentioned above mostly focus on adjusting network architectures in the encoding phase and the inherent problem of CTC is untouched.

\subsection{Attn based Text Recognition}
%1.Attn介绍,2.基于attn的方法介绍
%开山之作
%readed_2016_Recursive Recurrent Nets With Attention Modeling for OCR in the Wild
%下面几篇一起讲
%类似之作，加了rectification
%readed_2018_ASTER_ An Attentional Scene Text Recognizer with Flexible Rectification
%readed_2016_Robust Scene Text Recognition With Automatic Rectification
%readed_201901_MORAN- A Multi-Object Rectified Attention Network for Scene Text Recognition
%readed_201908_Symmetry-constrained Rectification Network for Scene Text Recognition
%其他方式处理2D问题
%readed_2017_Arbitrarily-Oriented Text Recognition
%readed_2017_Focusing Attention- Towards Accurate Text Recognition in Natural Images
%readed_2018.11_Show, attend and read_ a simple and strong baseline for recognizing irregular text
%seq2seq改成transformer
%readed_201904_A Simple and Robust Convolutional-Attention Network for Irregular Text Recognition
As to the methods based on Attn, they can be traced back to the work in ~\cite{Lee2016R2AM}. Inspired by the research in image captioning, five variations of RNNs are proposed in this paper and the model called R2AM outperforms the others, which can be considered as the prototype of subsequent Attn based approaches. Then some genres emerge. In ~\cite{Shi2016rar,shi2019aster,luo2019moran,Yang2019SCRN}, various rectification modules are applied to transform irregular text in images to regular ones. Besides, a fractional pickup strategy is designed to improve attention sensitivity in ~\cite{luo2019moran}. While some other researchers concentrate on boosting the attention mechanisms. Cheng \etal.~\cite{Cheng2017FAN} propose the Focusing Attention Network (FAN) to alleviate attention shifts and acquires more accurate position predictions. In ~\cite{Cheng2018AON}, to tackle with arbitrarily-oriented text images, feature maps from four directions are extracted and merged, then fed into Attn as attention maps. To address the irregular text problem, Li \etal.~\cite{li2019sar} employ a 2-dimensional attention map in the Attn based decoder. Recently, to make the most of attention mechanisms and reduce the training time cost, transformer-like ~\cite{VaswaniS2017transformer} structures are introduced in ~\cite{wang2019src} to replace RNNs in the Attn.

\begin{figure}[!t]
	\centering
	\includegraphics[width=3in]{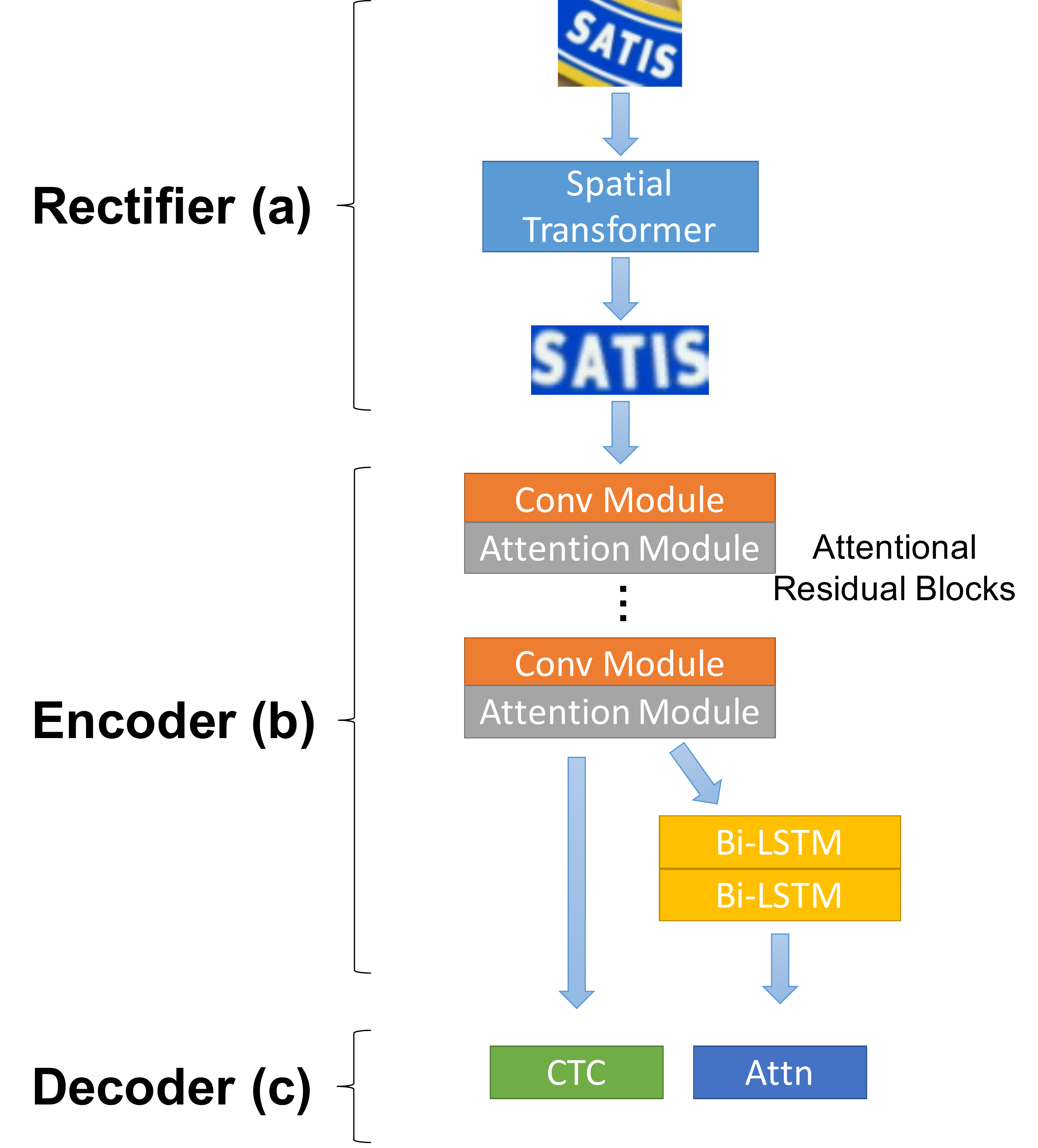}
	% where an .eps filename suffix will be assumed under latex,
	% and a .pdf suffix will be assumed for pdflatex; or what has been declared
	% via \DeclareGraphicsExtensions.
	\caption{Overview of the ReADS. (a) Rectifier using STN. (b) Encoder with stacked attentional residual blocks and Bi-LSTMs in two branches. (c) Decoding in two ways, namely CTC and attentional sequence recognition.}
	\label{framework}
\end{figure}

\section{Method}
The ReADS is composed of three parts, the rectifier, the encoder, and the decoder. An overview of the whole network architecture is provided in Figure~\ref{framework}. In detail, the rectifier is a light-weighted STN which adjusts the input images into more readable images of the same size. Next, the encoder with two branches takes rectified images and outputs two types of representations. Finally, these two outputs are decoded by CTC and Attn separately in the decoding phase. 

\subsection{Rectifier}
Since the Thin-Plate-Spline (TPS)~\cite{warps1989thin} proves to be more effective on perspective and curved text images than the simple affine transformation, we adopt an STN with a predicted TPS in this stage which makes the rectifier learnable. The STN consists of three parts which are the localization network, the grid generator and the sampler. First a set of control points is predicted by the localization network. Then the grid generator obtains a TPS transformation via the control points and outputs a sampling grid. Finally, the rectified images are sampled from the original images by the sampler.
 
\subsubsection{Localization Network}
A TPS transformation is calculated from two sets of control points with equal size ${K}$, denoted by $\mathbf{C}^s$ and $\mathbf{C}^t$ respectively. Here $\mathbf{C}^s=[c^s_1,...,c^s_K] \in \mathbb{R}^{2 \times K}$ is the list of source control points,  where $c^s_K = [x^s_k,y^s_k]^T$ is the $k$-th point.   $\mathbf{C}^t$ is the target control points which is similar to $\mathbf{C}^s$. 
The $\mathbf{C}^t$ are placed evenly along the top and bottom borders of the output image at fixed positions. Thus we only need to obtain $\mathbf{C}^s$  and it is acquired by the localization network. This network processes the input image via convolutional and pooling layers, then regresses $\mathbf{C}^s$ by a fully-connected layer whose output size is $2K$. It should be noted that the whole rectifier is differentiable and can be trained by the back-propagated gradients.

\subsubsection{Grid Generator}
Given $\mathbf{C}^s$ and $\mathbf{C}^t$, the gird generator computes a 2D TPS transformation and uses it to generate a sampling grid that maps every location in the rectified image to the input image.  The function of this transformation is shown as follows,
\begin{equation} 
\mathbf{C}^s=\textbf{a}+\mathbf{b}^T\mathbf{C}^t+\mathbf{w}^T\varphi(\mathbf{C}^t),  \label{eq_1}
\end{equation} 
where $\mathbf{a} \in \mathbb{R}^{2 \times 1}$, $\mathbf{b} \in \mathbb{R}^{2 \times 2}$, $\mathbf{w} \in \mathbb{R}^{K \times 2}$ and $\varphi$ is a function which can be calculated by these formations:
\begin{equation}
\begin{aligned}
 \varphi(\mathbf{C}^t)&=(\delta(\mathbf{C}^t-c^t_1),\delta(\mathbf{C}^t-c^t_2),...,\delta(\mathbf{C}^t-c^t_K))^T,  \\
 \delta(\mathbf{x})&=\left \| \mathbf{x} \right \|^2\log(\left \| \mathbf{x}  \right \|).  \label{eq_2}
\end{aligned}
\end{equation} 
By solving a linear system with boundary conditions,
\begin{equation}
\begin{aligned}
\mathbf{w}&=0,\\
\mathbf{C}_{x}^t \mathbf{w}_{1}&=0,\\
\mathbf{C}_{y}^t \mathbf{w}_{2}&=0, \label{eq_3}
\end{aligned}
\end{equation}
 the coefficients $\mathbf{a}$,$\mathbf{b}$ and $\mathbf{w}$ can be found. $\mathbf{w}_{1}$ and $\mathbf{w}_{2}$ are the first and second column of $\mathbf{w}$. $\mathbf{C}_{x}^t$ and $\mathbf{C}_{y}^t$ are the $x$ and $y$ coordinates of $\mathbf{C}^t$. Then the Eq.~\eqref{eq_1} can be rewritten as
\begin{equation}
\begin{bmatrix}
\textbf{w}\\ 
\textbf{a}\\
\textbf{b}
\end{bmatrix}= 
\begin{bmatrix}
\mathbf{S}& \mathbf{1}^{1 \times K} &\mathbf{C}^t \\ 
\mathbf{1}^{K \times 1}& 0 &0 \\ 
(\mathbf{C}^t)^{T}& 0 &0 
\end{bmatrix}^{-1}\begin{bmatrix}
\mathbf{C^s}\\ 
0\\ 
0
\end{bmatrix},
\label{eq_4}
\end{equation}
where $\mathbf{S} \in \mathbb{R}^{K \times K}$ is a matrix and $s_{i,j}=\delta(\left \| \mathbf{c}^t_i-\mathbf{c}^t_j \right \|)$.
\subsubsection{Sampler}
The sampler obtains every pixel value in the rectified image by interpolating in the input image. When the location is outside the input image, its value is clipped to keep inside the image. A bilinear interpolation strategy is applied which computes pixel values in the rectified image from the four nearest sampling pixels. Same as the localization network, the sampler is totally differentiable and this allows the rectifier to be optimized by gradients based algorithms.

\subsection{Encoder}
Previous methods mainly employ classic CNN structures  (\textit{e.g.}, VGG~\cite{simonyan2015vgg}, ResNet~\cite{He_2016_resnet} and InceptionNet~\cite{Szegedy_2015_inception}) as visual feature extractors in the encoder. However, there are various disturbances in scene text images. So we introduce attention mechanisms into the encoder design which can suppress invalid backgrounds and highlight the useful foregrounds. Constrained by the receptive fields in convolutional layers, RNNs are often utilized to enlarge context regions after the visual feature extractor. However, the local texture representation will be impaired after RNNs. We elaborate on two branches to extract features from different receptive field scales. The first branch directly passes the visual feature map outputted by CNNs to the decoder. The second branch processes the visual feature map by stacked Bi-LSTMs before feeding it into the decoder.
%之前的方法主要使用一些vgg,resnet,inception的作为encoder，由于scene text存在大量的背景干扰，我们考虑将attention机制引入其中，用于抑制背景，提取有效前景。
%基于RNN上下文编码引入了更多的上下文信息，扩大了感受野，但相应的local的纹理信息也会被削弱，因此我们设计了两个分支，一个直接将cnn输出的feature map 作为decoder的输入，另一个通过stacked bi-listm，得到带有上下文信息的feature sequence

\subsubsection{Attentional Residual Block}
\begin{figure}[!t]
	\centering
	\includegraphics[width=2.0in]{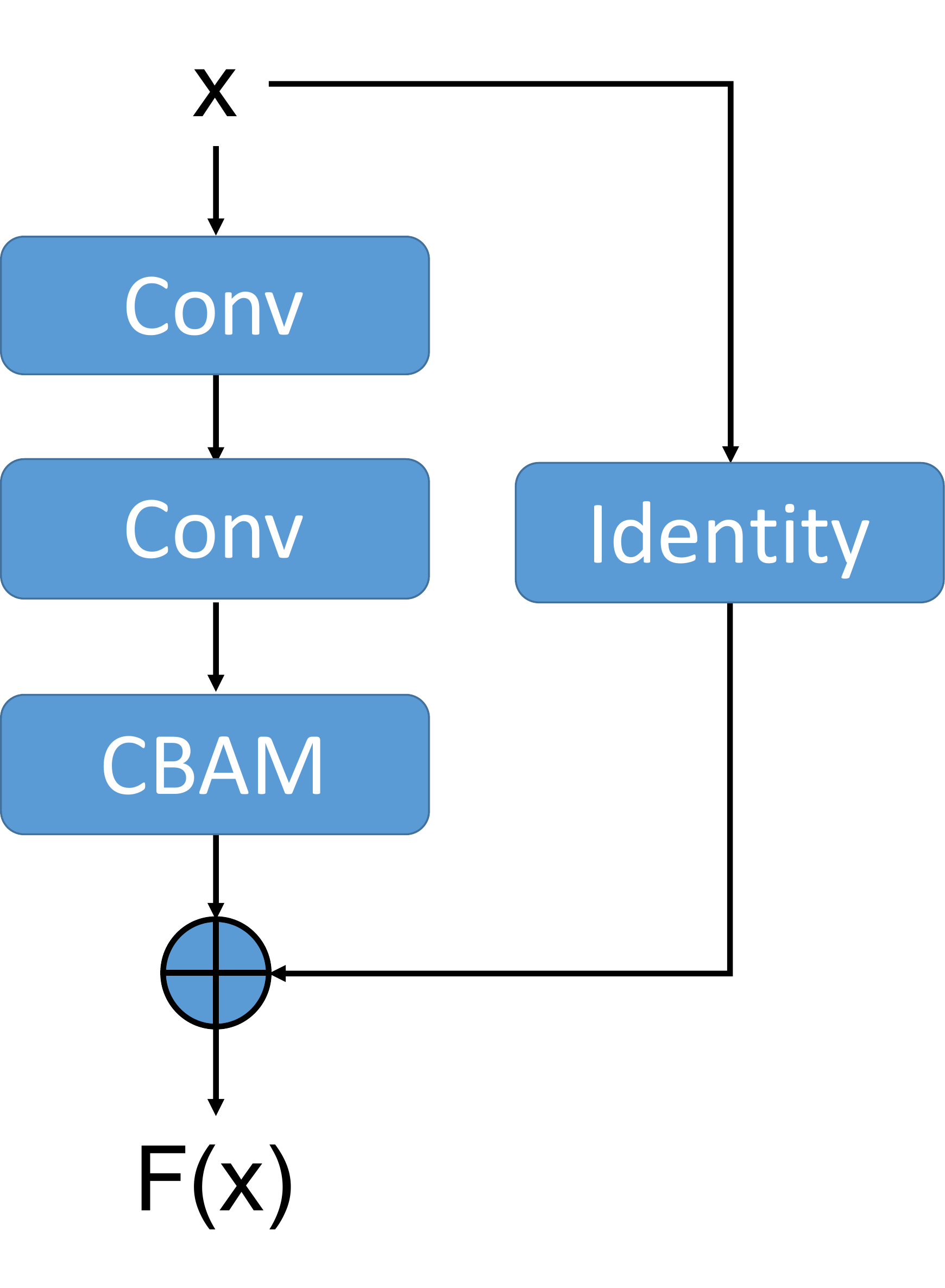}
	\caption{The structure of the attentional residual block.}
	\label{attention_res}
\end{figure}

In the encoding stage, we adopt attentional residual blocks to extract visual features. The structure of an attentional residual block is depicted in Figure~\ref{attention_res}. An attention mechanism called CBAM~\cite{Woo_2018_cbam} is implemented before merging the trunks and shortcuts. The CBAM learns channel-wise and spatial attentions separately with negligible parameter overheads. Moreover, it can be plugged into any residual blocks. 

The CBAM consists of two attention modules: the channel attention and spatial attention. Given an intermediate feature map $F$, firstly an average pooling operation $F_{avg}^c$ and a max-pooling operation $F_{max}^c$ are computed simultaneously to describe the global distinctive features. Then the channel attention mask $M_c(F) \in \mathbb{R}^{1 \times 1 \times C}$ is forwarded by a set of shared convolutional layers with both descriptors:

\begin{equation}
M_c(F)=sigmoid(\textit{Conv}(F_{avg}^c)+\textit{Conv}(F_{max}^c)).\label{eq_5}
\end{equation} 

Similarly to $M_c(F)$, two single-channel feature maps are obtained by channel-wise max-pooling and average-pooling. Then they are concatenated and processed by a  convolutional layer to acquire the spatial attention mask $M_a(F) \in \mathbb{R}^{W \times H \times 1}$:
\begin{equation}
M_a(F)=sigmoid(f^{3\times3}([F_{avg}^s;F_{max}^s])).\label{eq_6}
\end{equation}
Finally $F$ is broadcast multiplied by $M_c(F)$ and $M_a(F)$ to produce the attentional feature map.

\subsubsection{Bi-LSTM}
Long Short-Term Memory (LSTM) is capable of modeling long-range dependencies of the input sequential features. The output of LSTM at every timestamp depends not only on the current input but also on the previous inputs. Since the visual feature map is undirectional and the ability of single layer LSTM is limited, we stack two bi-directional LSTMs for context modeling. Given an input ${F}' \in \mathbb{R}^{{W}' \times {H}' \times {C}'}$, its size is reshaped to $({W}'{H}') \times {C}'$ before being sent into the stacked Bi-LSTMs. Then the size of output ${H}'$ is $({W}'{H}') \times {D}'$, where ${D}'$ is the number of hidden units.

\subsection{Decoder}
%scene text识别主要依赖于两方面，一个是图像中文字本身的特征，一个是文字之间的上下文依赖关系。为了从这两方面建模，并且利用他们的各自的优势，我们在decode的时候使用了两种技术。其中CTC主要负责利用文字本身的特征，因此使用encoder中的cnn特征作为输入。Attn则着重上下文关系建模，因此使用encoder中的RNN特征作为输入。
Scene text recognition mainly relies on two types of representation which are inherent texture features in text images and semantic context dependencies between characters. To model from these two aspects and take both advantages of them, we adopt two kinds of techniques in the decoding phase, namely CTC and Attn. The CTC is responsible for recognition using inherent texture features, hence it takes the visual feature map from the attentional CNNs of the encoder. While Attn mainly focuses on semantic context features, it utilizes the output from the stacked Bi-LSTMs in the encoder. Then these two losses ($L_{CTC}$ and $L_{Attn}$)  are weighted added for back-propagation in training. The total loss $L_{total} $ are calculated as follows,
\begin{equation}
L_{total} = L_{Attn} + \lambda L_{CTC}, \label{eq_7}
\end{equation} 
where $\lambda$ is a hyperparameter and set to $0.1$ empirically in our experiments.

\subsubsection{CTC}
There are many advantages of CTC, such as parallel training and parameter-free decoding. For scene text recognition, CTC allows the network
to select the most probable character sequence. The CTC output dimension is the class number plus one `blank' symbol denoted by $N+1$. Given an input sequence feature $x$ of length $T$, the probabilities of all possible ways in aligning all possible label sequences are outputted. Knowing that one label sequence $l$ can be represented by different alignments, the conditional probability distribution over $l$ is a summation of total probabilities in all possible alignments $\pi$. Then the probability $p(l|x)$ of a given label $l$ conditioned on $x$ can be calculated as follows,
\begin{equation}
\begin{aligned}
p(\pi|x)&=\prod_{t=1}^{T} y_t, \forall \pi \in (N+1)^T, \\
p(l|x)&=\sum_{\pi \in B^{-1}(l)}p(\pi|x), \label{eq_8}
\end{aligned}
\end{equation} 
where $y_t$ is the probability at time $t$ and $B^{-1}(l)$ denotes the set of sequences which are mapped to $l$ by $B$. In inference, the most probable labeling for the input sequence is selected:
\begin{equation}
h(x) = \arg \max_{l \le T}  p(l|x). \label{eq_9}
\end{equation}
Since the solution space is exponential to $T$, beam search or greedy decoding is applicable in predicting. We just choose the latter and detailed reasons will be given in the inference section.
%We just choose the greedy search as it is faster than the other one and differences between two results are negligible in our experiments.
\subsubsection{Attn}
The Attention-based sequence prediction (Attn) can also translate feature sequences to character sequences in arbitrary lengths but via a different mechanism. Attn not only takes the visual feature into account but also models output dependencies. Such a model is appealing due to its simplicity and powerfulness in sequence modeling and its ability to capture output dependencies in a recurrent way. 

Attn makes use of the encoder output at every decoding step by the attention mechanism. It proceeds iteratively for $M$ steps to generate a symbol sequence $\mathbf{Y} = (y_1,...,y_M)$ of length $M$ until an \textbf{E}nd \textbf{O}f \textbf{S}equence (\textit{EOS}) symbol. At step $m$, based on the output of the Bi-LSTMs encoder branch $\mathbf{H} = (h1,...,h_T)$, the $y_m$ is predicted using the following formulations,
\begin{equation}
\begin{aligned}
y_m & = softmax(W_o s_m + b_o), \\
s_m & = LSTM(y_{m-1},s_{m-1},c_m), \\
c_m & = \sum_{i=1}^{T}\alpha_{mi}h_{i}, \\
\alpha_{mi} & = \frac{\exp(e_{mi})}{\exp (\sum_{j=1}^{T}e_{mj})}, \\
e_{mi} &= v^T \tanh(Ws_{m-1}+Vh_i+b),
\end{aligned}
\end{equation}

where $W_o,b_o,v,W,V$ and $b$ are all learnable parameters and $s_m$ is the hidden state in the LSTM decoder at step $m$. $e_{mi}$ is an alignment model which scores how well the inputs around position $m$ and the output at position $i$ match.

\subsection{Inference}
During inference, there are two predictions outputted by two branches separately. We choose the character sequence from Attn as the final result because the accuracy of the Attn branch is always better than that of CTC branch under any experimental conditions. We leave the problem of merging predictions from two branches as a future research topic. As for the decoding method, different from most previous studies, we simply use the greedy decoding in both branches because it is faster than beam search and the performance differences are negligible in our experiments.

\begin{table}[!t] 
	\renewcommand{\arraystretch}{1.5}
	\caption{Text recognition network configurations. Each block is an attentional residual block. `k',`s' and `c' are kernel size, stride and channel number respectively. `Pool' stands for the Max pooling. `*' means a variable length.}
	\label{config_table}
	\centering
	\begin{adjustbox}{width=0.5\textwidth,center}
		\begin{tabular}{c|c|c|c|c|}
			\hline
			\multirow{2}{*}{}        & Layers                                                  & \multicolumn{2}{c|}{Configurations}                                                                                        & Output \\ \cline{2-5} 
			& Input                                                   & \multicolumn{2}{c|}{RGB image}                                                                                             & $32 \times 100$ \\ \hline
			Rectifier                & Localizer                                               & \multicolumn{2}{c|}{\begin{tabular}[c]{@{}c@{}}Conv $\times 6$ \\ 
					k : $3 \times 3$, s : $2 \times 2$ \\
					c :    {[}$32,64,128,256,256,256${]}\\ FC $\times 2$ \\
					units : {[}$512,20${]}  \end{tabular}} & $32 \times 100$ \\ \hline
			\multirow{7}{*}{Encoder} & Conv                                                    & \multicolumn{2}{c|}{k : $3 \times 3$, c :  $32$}                                                                                    & $32 \times 100$ \\ \cline{2-5} 
			& Block1                                                  & \multicolumn{2}{c|}{\begin{tabular}[c]{@{}c@{}}
					$\begin{bmatrix}
					\rm Conv \\ 
					\rm Conv \\
					\rm CBAM
					\end{bmatrix}$ $\times$ 3, k : $3 \times 3$, c :  $64$
					\\ Pool k : $2 \times 2$, s : $2 \times 2$ 
			\end{tabular}}                           & $16 \times 50$  \\ \cline{2-5} 
			& Block2                                                  & \multicolumn{2}{c|}{\begin{tabular}[c]{@{}c@{}}
					$\begin{bmatrix}
					\rm Conv \\ 
					\rm Conv \\
					\rm CBAM
					\end{bmatrix}$ $\times$ 4, k : $3 \times 3$, c :  $128$
					\\ Pool k : $2 \times 1$, s : $2 \times 2$ 
			\end{tabular}}                           & $8 \times 50$   \\ \cline{2-5} 
			& Block3                                                  & \multicolumn{2}{c|}{\begin{tabular}[c]{@{}c@{}}
					$\begin{bmatrix}
					\rm Conv \\ 
					\rm Conv \\
					\rm CBAM
					\end{bmatrix}$ $\times$ 6, k : $3 \times 3$, c :  $256$
					\\ Pool k : $2 \times 1$, s : $2 \times 2$ 
			\end{tabular}}                          & $4 \times 50$   \\ \cline{2-5} 
			& Block4                                                  & \multicolumn{2}{c|}{\begin{tabular}[c]{@{}c@{}}
					$\begin{bmatrix}
					\rm Conv \\ 
					\rm Conv \\
					\rm CBAM
					\end{bmatrix}$ $\times$ 3, k : $3 \times 3$, c :  $512$
					\\ Pool k : $2 \times 1$, s : $2 \times 2$ 
			\end{tabular}}                          & $2 \times 50$   \\ \cline{2-5} 
			& Conv                                                    & \multicolumn{2}{c|}{k : $2 \times 2$, c : $1024$}                                                                                  & $1 \times 49$   \\ \cline{2-5} 
			& \begin{tabular}[c]{@{}c@{}}Attn and CTC \\ Branches\end{tabular} & \begin{tabular}[c]{@{}c@{}} {[}Bi-LSTM{]} $\times$ 2  \\ hidden units : {[}512,512{]} \end{tabular}                      & -                     & $49$     \\ \hline
			Decoder                  & \begin{tabular}[c]{@{}c@{}}Attn and CTC \\ Branches\end{tabular} & \begin{tabular}[c]{@{}c@{}}Attentional LSTM\\  hidden units : 1024 \\
				attention units:1024 \end{tabular}                   & CTC                   & *      \\ \hline
		\end{tabular}
	\end{adjustbox}
\end{table}

\begin{table*}[!t] 
	\renewcommand{\arraystretch}{1.5}
	\caption{Results of our model compared with other proposed models.Numbers in bold indicate the best and numbers with underline depict second-best performance. Only lexicon-free results are reported.}
	\label{sota}
	\centering
	\begin{adjustbox}{width=1.0\textwidth,center}
		\begin{tabular}{l|l|l|l|l|l|l|l|l}
			\hline
			\multicolumn{1}{c|}{Method}                     & \multicolumn{4}{c|}{Regular Text}                                                                               & \multicolumn{4}{c}{Irregular Text}                                                                                    \\ \hline
			\multicolumn{1}{r|}{}   & \multicolumn{1}{c|}{IIIT5K} & \multicolumn{1}{c|}{SVT}  & \multicolumn{1}{c|}{IC03} & \multicolumn{1}{c|}{IC13} & \multicolumn{1}{c|}{IC15-2077} & \multicolumn{1}{c|}{IC15-1811} & \multicolumn{1}{c|}{SVTP} & \multicolumn{1}{c}{CUTE} \\ \hline \hline
			Jaderberg \etal. 2014~\cite{corr2014Jaderberg} & \multicolumn{1}{c|}{-}      & \multicolumn{1}{c|}{80.7} & \multicolumn{1}{c|}{93.1}    & \multicolumn{1}{c|}{90.8} & \multicolumn{1}{c|}{-}         & \multicolumn{1}{c|}{-}         & \multicolumn{1}{c|}{-}    & \multicolumn{1}{c}{-}    \\ \hline
			Shi \etal. 2016~\cite{shi2017crnn}                                   & \multicolumn{1}{c|}{78.2}   & \multicolumn{1}{c|}{80.8} & \multicolumn{1}{c|}{89.4}    & \multicolumn{1}{c|}{86.7} & \multicolumn{1}{c|}{-}         & \multicolumn{1}{c|}{-}         & \multicolumn{1}{c|}{-}    & \multicolumn{1}{c}{-}    \\ \hline
			Shi \etal. 2016~\cite{Shi2016rar}                                  & \multicolumn{1}{c|}{81.9}       & \multicolumn{1}{c|}{81.9}     & \multicolumn{1}{c|}{90.1}     & \multicolumn{1}{c|}{88.6}     & \multicolumn{1}{c|}{-}          & \multicolumn{1}{c|}{-}          & \multicolumn{1}{c|}{71.8}     & \multicolumn{1}{c}{59.2}     \\ \hline
			Liu \etal. 2016~\cite{liu2016starnet}                                       & \multicolumn{1}{c|}{83.3}       & \multicolumn{1}{c|}{83.6}     & \multicolumn{1}{c|}{89.9}     & \multicolumn{1}{c|}{89.1}     & \multicolumn{1}{c|}{-}          & \multicolumn{1}{c|}{-}          & \multicolumn{1}{c|}{73.5}     & \multicolumn{1}{c}{-}     \\ \hline
			Gao \etal. 2017~\cite{Gao2017ACSM}		& \multicolumn{1}{c|}{81.8}       & \multicolumn{1}{c|}{82.7}     & \multicolumn{1}{c|}{89.2}     & \multicolumn{1}{c|}{88.0}     & \multicolumn{1}{c|}{-}          & \multicolumn{1}{c|}{-}          & \multicolumn{1}{c|}{-}     & \multicolumn{1}{c}{-}     \\ \hline
			Cheng \etal. 2018~\cite{Cheng2018AON}                              & \multicolumn{1}{c|}{87.0}       & \multicolumn{1}{c|}{82.8}     & \multicolumn{1}{c|}{91.5}     & \multicolumn{1}{c|}{-}     & \multicolumn{1}{c|}{68.2}          & \multicolumn{1}{c|}{-}          & \multicolumn{1}{c|}{73.0}     & \multicolumn{1}{c}{76.8}     \\ \hline
			Liu \etal. 2018~\cite{Liu2018CharNet} & \multicolumn{1}{c|}{83.6}       & \multicolumn{1}{c|}{84.4}     & \multicolumn{1}{c|}{91.5}     & \multicolumn{1}{c|}{90.8}     & \multicolumn{1}{c|}{60.0}          & \multicolumn{1}{c|}{-}          & \multicolumn{1}{c|}{73.5}     & \multicolumn{1}{c}{-}     \\ \hline
			Shi \etal. 2019~\cite{shi2019aster} & \multicolumn{1}{c|}{\underline{93.4}}       & \multicolumn{1}{c|}{\textbf{93.6}}     & \multicolumn{1}{c|}{94.5}     & \multicolumn{1}{c|}{91.8}     & \multicolumn{1}{c|}{-}          & \multicolumn{1}{c|}{76.1}          & \multicolumn{1}{c|}{78.5}     & \multicolumn{1}{c}{79.5}     \\ \hline
			Liao \etal. 2019~\cite{liao2019scene} &  \multicolumn{1}{c|}{92.0}     &        \multicolumn{1}{c|}{82.1}          &          \multicolumn{1}{c|}{-}    &        \multicolumn{1}{c|}{91.4}        &         \multicolumn{1}{c|}{-}              &         \multicolumn{1}{c|}{-}           &   \multicolumn{1}{c|}{-}              &      \multicolumn{1}{c}{78.1}         \\ \hline
			Zhan \& Lu \etal. 2019~\cite{zhan2019esir} &    \multicolumn{1}{c|}{93.3}      &      \multicolumn{1}{c|}{90.2}           &   \multicolumn{1}{c|}{-}        &       \multicolumn{1}{c|}{91.3}        &      \multicolumn{1}{c|}{-}        &    \multicolumn{1}{c|}{76.9}     &    \multicolumn{1}{c|}{\underline{79.6}}    &   \multicolumn{1}{c}{\underline{83.3}}     \\ \hline
			Luo \etal. 2019~\cite{luo2019moran}  &     \multicolumn{1}{c|}{91.2}    &     \multicolumn{1}{c|}{88.3}        &    \multicolumn{1}{c|}{\underline{95.0}}  &      \multicolumn{1}{c|}{92.4}        &        \multicolumn{1}{c|}{68.8}       &        \multicolumn{1}{c|}{-}          &        \multicolumn{1}{c|}{76.1}         &     \multicolumn{1}{c}{77.4}      \\ \hline
			Gao \etal. 2019~\cite{gao12text} &     \multicolumn{1}{c|}{89.9}    &     \multicolumn{1}{c|}{87.2}        &    \multicolumn{1}{c|}{93.3}  &      \multicolumn{1}{c|}{92.9}        &        \multicolumn{1}{c|}{\underline{74.5}}       &        \multicolumn{1}{c|}{-}          &        \multicolumn{1}{c|}{76.4}         &     \multicolumn{1}{c}{70.8}      \\ \hline
			Baek \etal. 2019~\cite{Baek_2019_what} &     \multicolumn{1}{c|}{87.9}    &     \multicolumn{1}{c|}{87.5}        &    \multicolumn{1}{c|}{94.4}  &      \multicolumn{1}{c|}{92.3}        &        \multicolumn{1}{c|}{71.8}       &        \multicolumn{1}{c|}{\underline{77.6}}          &        \multicolumn{1}{c|}{79.2}         &     \multicolumn{1}{c}{74.0}      \\ \hline
			Liu \etal. 2019~\cite{liu2019safe} &     \multicolumn{1}{c|}{85.2}    &     \multicolumn{1}{c|}{85.5}        &    \multicolumn{1}{c|}{92.9}  &      \multicolumn{1}{c|}{90.3}        &        \multicolumn{1}{c|}{65.7}       &        \multicolumn{1}{c|}{71.8}          &        \multicolumn{1}{c|}{74.4}         &     \multicolumn{1}{c}{-}      \\ 
			\hline
			Wan \etal. 2020~\cite{wan20192dctc} &     \multicolumn{1}{c|}{\textbf{94.7}}    &     \multicolumn{1}{c|}{90.6}        &    \multicolumn{1}{c|}{-}  &      \multicolumn{1}{c|}{\underline{93.9}}        &        \multicolumn{1}{c|}{-}       &        \multicolumn{1}{c|}{75.2}          &        \multicolumn{1}{c|}{79.2}         &     \multicolumn{1}{c}{81.3}      \\
			\hline
			Wang \etal. 2020~\cite{wang2020faclstm} &     \multicolumn{1}{c|}{90.5}    &     \multicolumn{1}{c|}{82.2}        &    \multicolumn{1}{c|}{-}  &      \multicolumn{1}{c|}{-}        &        \multicolumn{1}{c|}{-}       &        \multicolumn{1}{c|}{-}          &        \multicolumn{1}{c|}{-}         &     \multicolumn{1}{c}{\underline{83.3}}      \\
			 \hline  \hline
			\textbf{Ours}  &     \multicolumn{1}{c|}{91.0}    &     \multicolumn{1}{c|}{\underline{91.2}}        &    \multicolumn{1}{c|}{\textbf{96.1}}  &      \multicolumn{1}{c|}{\textbf{94.5}}        &        \multicolumn{1}{c|}{\textbf{75.1}}       &        \multicolumn{1}{c|}{\textbf{80.4}}          &        \multicolumn{1}{c|}{\textbf{83.3}}         &     \multicolumn{1}{c}{\textbf{83.7}}      \\ \hline
		\end{tabular}
	\end{adjustbox}
\end{table*}

\section{Experiments}
In this section, we conduct extensive experiments on various scene text recognition benchmarks to verify the effectiveness of our model. First the datasets we perform experiments on are introduced in Section~\ref{dataset}. Then the model implementation details in training and inferring are described in Section~\ref{impl}. Finally, we analyze and demonstrate the results in comparing experiments with other state-of-the-art methods and some ablation studies can be found in Section~\ref{exp}.

\subsection{Datasets}{\label{dataset}}
The ReADS is only trained on two public synthetic datasets, MJSynth~\cite{corr2014Jaderberg} and SynthText~\cite{Gupta2016CVPR}, without finetuning on other datasets. Then the model is tested on 7 datasets, which are ICDAR 2003 (IC03)~\cite{ic03} , ICDAR 2013 (IC13)~\cite{ic13}, ICDAR 2015 (IC15)~\cite{ic15}, IIIT5K-Words (IIIT5K)~\cite{IIIT5K}, Street View Text (SVT)~\cite{svt}, Street View Text Perspective (SVTP)~\cite{SVTP}, and CUTE80 (CUTE)~\cite{cute}.

\noindent
{\bf MJSynth~\cite{corr2014Jaderberg}} 
is a synthetic text dataset. The dataset consists of about 9 million images covering 90k English words and includes the training, validation and testing splits separately. Random transformations and other effects are applied to every word image. All the images in MJSynth are taken for our model training.

\noindent
{\bf SynthText~\cite{Gupta2016CVPR}} 
is also a synthetic text dataset. But unlike MJSynth, it is intended for scene text detection. Words are rendered on whole images and not cropped. We extract all the word regions by the given word bounding boxes for training.

\noindent 
{\bf ICDAR 2003 (IC03)~\cite{ic03}} 
contains 1156 images for training and 1110 images for evaluation. Following Wang \etal. ~\cite{svt}, words which are either too short (less than 3 characters) or contain non-alphanumeric characters are ignored. Then the number of images for evaluation reduces to 867 after filtering.

\noindent
{\bf ICDAR 2013 (IC13)~\cite{ic13}} 
inherits most images from IC03 and adds some new ones. The trainset and testset have 848 and 1095 images separately. After removing words with non-alphanumeric characters, the filtered testset contains 1015 images for our evaluation.

\noindent
{\bf ICDAR 2015 (IC15)~\cite{ic15}} 
consists of 4468 images for training and 2077 images for evaluation. All the images are acquired by a pair of Google Glasses without deliberate positioning and focusing. Thus this dataset contains many noisy, blurry and irregular text. There are two versions of testset with different image numbers: 1811 and 2077. The former filters images with non-alphanumeric characters, extremely transformation and curved text out, while the later keeps all the images. We test our ReADS on both of the two versions.

\noindent
{\bf IIIT5K-Words (IIIT5K)~\cite{IIIT5K}} 
contains a trainset of 2000 images and a testset of 3000 images gathered from the Internet. Each image is associated with a 50-word lexicon and a 1,000-word lexicon.

\noindent
{\bf Street View Text (SVT)~\cite{svt}} 
consists of 249 images collected from Google Street View. The testset contains 647 cropped samples which are collected from these images. Many of the images are very noisy or have very low resolutions.

\noindent
{\bf Street View Text Perspective (SVTP)~\cite{SVTP}} 
is a collection of 645 images from Google Street View like SVT. But most of the images are more difficult to recognize because of large perspective distortions.

\noindent
{\bf CUTE80 (CUTE)~\cite{cute}} 
is a dataset with 288 cropped images and most images in it contains curved text. It is collected from natural scenes.

\subsection{Implementation Details}{\label{impl}}
The network configurations are summarized in Table~\ref{config_table}. A 34-layer residual network with CBAM is adopted as the visual feature extractor. Except for the first residual block, each residual block is followed by an asymmetric max pooling to keep more horizontal resolution. Two $3 \times 3$ convolutions and a CBAM constitute every residual unit. Following the visual feature extractor, the network splits into two branches. In the CTC branch, the visual feature map is directly sent into the decoder for recognition. In the Attn branch, a stacked RNN of two layer Bi-LSTMs is in front of the attentional LSTM based decoder. Both the Attn and CTC recognize 62 classes, including digits, uppercase and lowercase letters. When evaluating the trained model on benchmarks, we normalize the predictions to case-insensitive and discard punctuation marks . Moreover, no lexicons are applied after predicting.

The proposed model is implemented by Tensorflow and is trained from scratch. The ADADELTA~\cite{zeiler2012adadelta} is applied as the optimizer with a batch size of 64. The total training iterations are about 4.5 epochs and images of every batch are randomly selected from MJSynth and SynthText. The initial learning rate is set to 1.0, then adjusted to 0.1 and 0.01 at the end of the 3rd and 4th epoch. All experiments are conducted on a workstation with a 2.20G Hz Intel(R) Xeon(R) CPU, 256GB RAM and two NVIDIA Tesla V100 GPUs. Some image processes such as random rotation ($[-3^{\circ},+3^{\circ}] $), elastic deformation, hue, brightness and contrast are applied for data augmentation in training. For images whose height are three times larger than width, we simply rotate the images anticlockwise by $90^{\circ}$. It should be noted that unlike some previous work, we only use greedy search and no input augmentations or results merging is adopted for inference.

\subsection{Results}{\label{exp}}
\subsubsection{Comparison with State-of-the-art}
In this section, we compare the ReADS with several state-of-the-art methods on the benchmarks mentioned in Section 4.1, as shown in Table ~\ref{sota}. Since our model is trained on synthetic data and does not use the model ensemble and results merging, we only list results acquired in the same condition for a fair comparison. It can be observed that, on both regular and irregular datasets, the ReADS can achieve either highly competitive or state-of-the-art performance. In summary, our method gets five first, one second and one competitive results on a total of seven benchmarks. Especially on IC03 and IC13, we outperform the previous methods by a relatively large margin. Our results on the two regular text datasets, IIIT5K and SVT, are slightly worse than results on IC03 and IC13. Through our observation and analysis, a possible explanation is that data sources of these two are relatively limited. Hence, the semantic context in the Attn branch is more important and the CTC branch does not fully take effect. Results of the ReADS on three irregular text datasets fluctuate by a certain degree. According to our analysis, there are two main reasons. The first is that no other mechanisms are introduced to tackle irregular text except the rectifier. However, the rectifier works in a weakly supervised way (only supervised by the recognition loss), so the effect is not very ideal in some difficult scenarios. The second reason is the data distributions and scales of these datasets are rather different. IC15 has the largest data scale and includes horizontal, Inclined and curved text images. SVTP is a medium scale dataset and consists of only horizontal and Inclined text images. The images in CUTE are the fewest and most of them are curved text images. 

%在规则和不规则数据集上，都取得了很有竞争力或sota的结果。
%尤其在几个经典的规则ic03，ic13，数据集上，而在iit和svt上效果稍差，经过我们的观察和分析，一个可能的解释是，这两个数据集的来源比较单一，因此语义性比较强，CTC分支的效果不明显。在三个不规则数据上效果有一定起伏，我们分析主要有两个原因，第一是除了tps，我们并没有引入其他的机制用来处理不规则文字，而tps这种基于间接监督（最终识别loss）信号的校正模块，处理非常不规则的文字行时，效果不是很理想。第二是这几个数据集的规模和分布差异较大，ic15数据较多，包含近似水平&倾斜&弯曲文字行，svtp数据中等，包含水平和倾斜，而cute数据最少且大部分都是弯曲文字

%Figure 8: Some bad cases produced by our recognition sys- tem. The meanings of these elements are the same as Fig. 6.
%Incorrectly recognized characters are in red.

\subsubsection{Ablation Study}
In this section, we conduct two sets of experiments for ablation studies. The first is to analyze the impact of some modules in the network. The second is to verify the effectiveness of double supervised branches.

\noindent
{\bf Influence of Modules:} In this part, we perform several experiments to explore the impact of some modules on the model performance. Two key modules are taken into consideration, the rectifier and the attention mechanisms. All the implementation details are the same as those of the last section and results are demonstrated in Table \ref{ab_modules}. We observe that the rectifier is effective for the irregular text while a little harmful for regular text. Another conclusion is that the attention mechanisms achieve consistent improvement on all the benchmarks except CUTE. We believe that it is due to the small scale of CUTE. There are only 288 images in CUTE, so one more wrong recognized text image brings about $0.35\%$ reduction on the accuracy. Fortunately, by introducing both two modules, we can take all the advantages from them and obtain better performance than other settings.

\begin{table}[!t] 
	\renewcommand{\arraystretch}{1.5}
	\caption{Results of using different modules. Models without attentions means that all CBAM modules are removed from residual blocks. The best accuracies on benchmarks is in bold.}
	\label{ab_modules}
	\centering
	\begin{adjustbox}{width=0.5\textwidth,center}
	\begin{tabular}{|c|c|c|c|c|c|c|c|c|c|}
		\hline
		\multicolumn{2}{|c|}{Modules}   & \multicolumn{4}{c|}{Regular Text}                             & \multicolumn{4}{c|}{Irregular Text}                           \\ \hline
		Rectifier & Attentions & IIIT5K        & SVT           & IC03          & IC13          & IC15-2077     & IC15-1811     & SVTP          & CUTE          \\ \hline
		        &                   & 89.4          & 87.6          & 94.8          & 93.1          & 70.4          & 75.0          & 76.7          & 80.2          \\ \hline
		        & \checkmark                 & 90.0          & 90.0          & 95.3          & 93.4          & 74.3          & 79.2          & 80.3          & 77.4          \\ \hline
		\checkmark        &                   & 90.1          & 90.3          & 94.6          & 92.3          & 72.8          & 78.4          & 81.2          & \textbf{83.7}          \\ \hline
		\checkmark        & \checkmark                  & \textbf{91.0} & \textbf{91.2} & \textbf{96.1} & \textbf{94.5} & \textbf{75.1} & \textbf{80.4} & \textbf{83.3} & \textbf{83.7} \\ \hline
	\end{tabular}
\end{adjustbox}
\end{table}

\begin{figure}[!t]
	\centering
	\includegraphics[width=3in]{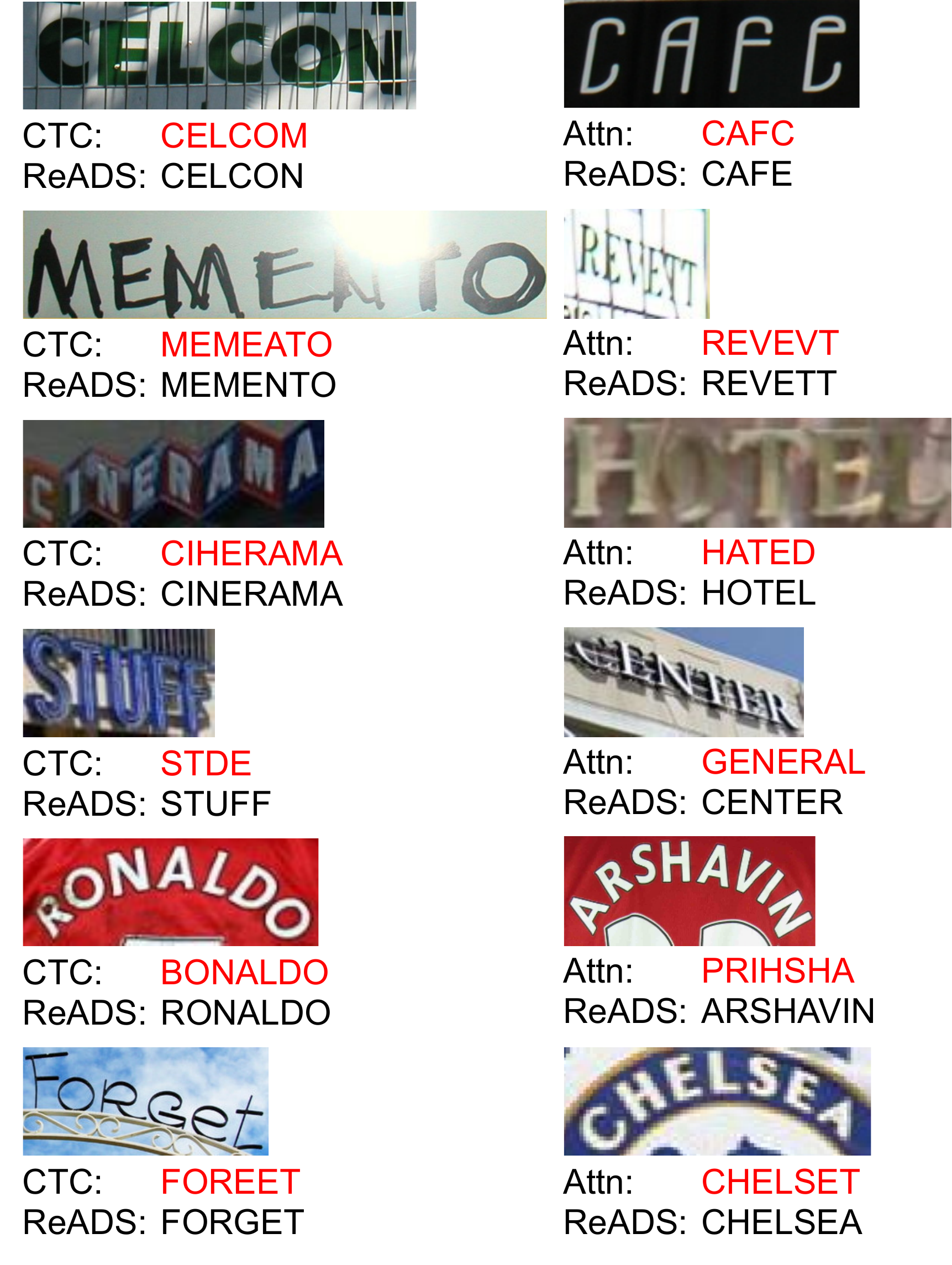}
	\caption{Some compared cases selected from the benchmarks. The images in the left column are recognized successful by the ReADS and failed by the CTC recognizer. The images in the right column are recognized successful by the ReADS and failed by the Attn recognizer.}
	\label{compare}
\end{figure}

%如图所示，尽管ctc在所有数据集上结果都比attn差，但它依然给rads的性能带来了提升。在大部分测试集上rads都超过了两种单分支方法。在不规则数据集上，attn和rads方法结果相差不大，我们认为这是由于缺乏更有效的处理不规则数据机制，而目前双分支监督的性能已经饱和造成的
\noindent
{\bf Influence of Branches:} Meanwhile, we design some experiments to verify the effectiveness of double branch supervising. We compare the ReADS with two implementations which disable one of the two supervised branches and keep other components unchanged. As is shown in Table \ref{ab_branches}, although the CTC branch network performs worse than the Attn branch network, it still provides effective supervision in the ReADS. The double branch supervised network outperforms any single branch supervised versions on most benchmarks. We observe that, on the irregular text datasets, the Attn branch network achieves comparable results with the ReADS. According to our analysis, it is due to the lack of more effective rectifying modules and the performance of double supervised branches is saturated for irregular text images. Some qualitative cases are also illustrated in Figure \ref{compare} to verify the effectiveness of the ReADS.

\begin{table}[!t] 
	\renewcommand{\arraystretch}{1.5}
	\caption{Results of using different supervised branches. Numbers in bold are the best performance.}
	\label{ab_branches}
	\centering
	\begin{adjustbox}{width=0.5\textwidth,center}
	\begin{tabular}{|c|c|c|c|c|c|c|c|c|c|}
		\hline
		\multicolumn{2}{|c|}{Branches} & \multicolumn{4}{c|}{Regular Text}                             & \multicolumn{4}{c|}{Irregular Text}                           \\ \hline
		Attn           & CTC           & IIIT5K        & SVT           & IC03          & IC13          & IC15-2077     & IC15-1811     & SVTP          & CUTE          \\ \hline
		             & \checkmark            & 88.6          & 87.3          & 92.4          & 90.3          & 72.1          & 76.5          & 77.1          & 78.8          \\ \hline
		\checkmark             &             & \textbf{91.0} & 90.6          & 94.3          & 93.3          & \textbf{75.7} & 80.2          & \textbf{84.2} & 82.3          \\ \hline
		\checkmark             & \checkmark            & \textbf{91.0} & \textbf{91.2} & \textbf{96.1} & \textbf{94.5} & 75.1          & \textbf{80.4} & 83.3          & \textbf{83.7} \\ \hline
	\end{tabular}
\end{adjustbox}
\end{table}

\section{Conclusion}
In this paper, we present a novel approach named ReADS, a rectified attentional double supervised scene text recognizer. It is equipped with decoders of the Attention based sequence recognition (Attn) and the Connectionist Temporal Classification (CTC) for semantic context modeling and inherent texture representation. Combined with two effective modules, the rectifier, and attention mechanisms, the ReADS shows state-of-the-art or highly competitive performance on seven benchmarks compared with previous works. Moreover, the model can be trained end-to-end from scratch and no additional labels are needed except the line-level text. In the future, we plan to explore stronger attention mechanisms especially for irregular text. Merging predictions from CTC and Attn branches is another interesting topic. Besides, since scene text recognition is very relevant to NLP, we will combine
these two techniques for better results.


%\section*{Acknowledgment}




\bibliographystyle{IEEEtran}
\bibliography{IEEEabrv,bare_conf_RADS_20191230}

\begin{thebibliography}{10}
\providecommand{\url}[1]{#1}
\csname url@samestyle\endcsname
\providecommand{\newblock}{\relax}
\providecommand{\bibinfo}[2]{#2}
\providecommand{\BIBentrySTDinterwordspacing}{\spaceskip=0pt\relax}
\providecommand{\BIBentryALTinterwordstretchfactor}{4}
\providecommand{\BIBentryALTinterwordspacing}{\spaceskip=\fontdimen2\font plus
\BIBentryALTinterwordstretchfactor\fontdimen3\font minus
  \fontdimen4\font\relax}
\providecommand{\BIBforeignlanguage}[2]{{%
\expandafter\ifx\csname l@#1\endcsname\relax
\typeout{** WARNING: IEEEtran.bst: No hyphenation pattern has been}%
\typeout{** loaded for the language `#1'. Using the pattern for}%
\typeout{** the default language instead.}%
\else
\language=\csname l@#1\endcsname
\fi
#2}}
\providecommand{\BIBdecl}{\relax}
\BIBdecl

\bibitem{shi2017crnn}
B.~Shi, X.~Bai, and C.~Yao, ``An end-to-end trainable neural network for
  image-based sequence recognition and its application to scene text
  recognition,'' \emph{IEEE Transactions on Pattern Analysis and Machine
  Intelligence}, vol.~39, no.~11, pp. 2298--2304, 2017.

\bibitem{liu2016starnet}
W.~Liu, C.~Chen, K.-Y.~K. Wong, Z.~Su, and J.~Han, ``Star-net: a spatial
  attention residue network for scene text recognition,'' in \emph{The British
  Machine Vision Conference (BMVC)}, 2016.

\bibitem{Gao2017ACSM}
Y.~Gao, Y.~Chen, J.~Wang, and H.~Lu, ``Reading scene text with attention
  convolutional sequence modeling,'' \emph{Neurocomputing}, 2017.

\bibitem{corr2014Jaderberg}
M.~Jaderberg, K.~Simonyan, A.~Vedaldi, and A.~Zisserman, ``Synthetic data and
  artificial neural networks for natural scene text recognition,'' \emph{CoRR},
  vol. abs/1406.2227, 2014.

\bibitem{Gupta2016CVPR}
A.~Gupta, A.~Vedaldi, and A.~Zisserman, ``Synthetic data for text localisation
  in natural images,'' in \emph{The IEEE Conference on Computer Vision and
  Pattern Recognition (CVPR)}, 2016.

\bibitem{NIPS2015stn}
M.~Jaderberg, K.~Simonyan, A.~Zisserman, and k.~kavukcuoglu, ``Spatial
  transformer networks,'' in \emph{Advances in Neural Information Processing
  Systems}, 2015, pp. 2017--2025.

\bibitem{Hochreiter1997lstm}
S.~Hochreiter and J.~Schmidhuber, ``Long short-term memory,'' \emph{Neural
  Computation}, vol.~9, no.~8, pp. 1735--1780, 1997.

\bibitem{c_c}
L.~Neumann and J.~Matas, ``Real-time scene text localization and recognition,''
  in \emph{IEEE Conference on Computer Vision and Pattern Recognition (CVPR)},
  2012.

\bibitem{Yao_2014_CVPR}
C.~Yao, X.~Bai, B.~Shi, and W.~Liu, ``Strokelets: A learned multi-scale
  representation for scene text recognition,'' in \emph{The IEEE Conference on
  Computer Vision and Pattern Recognition (CVPR)}, 2014.

\bibitem{15549WordSpotting}
K.~Wang and S.~Belongie, ``Word spotting in the wild,'' in \emph{ECCV}, 2010.

\bibitem{e2e}
K.~Wang, B.~Babenko, and S.~Belongie, ``End-to-end scene text recognition,'' in
  \emph{International Conference on Computer Vision (ICCV)}, 2011.

\bibitem{graves2006ctc}
A.~Graves, S.~Fern\'{a}ndez, F.~Gomez, and J.~Schmidhuber, ``Connectionist
  temporal classification: Labelling unsegmented sequence data with recurrent
  neural networks,'' in \emph{The 23rd International Conference on Machine
  Learning (ICML)}, 2006.

\bibitem{xie2016fcrn}
{Zecheng Xie}, {Zenghui Sun}, {Lianwen Jin}, {Ziyong Feng}, and {Shuye Zhang},
  ``Fully convolutional recurrent network for handwritten chinese text
  recognition,'' in \emph{The 23rd International Conference on Pattern
  Recognition (ICPR)}, 2016.

\bibitem{Rosetta}
F.~Borisyuk, A.~Gordo, and V.~Sivakumar, ``Rosetta: Large scale system for text
  detection and recognition in images,'' in \emph{International Conference on
  Knowledge Discovery \& Data Mining (SIGKDD)}, 2018.

\bibitem{Lee2016R2AM}
C.-Y. Lee and S.~Osindero, ``Recursive recurrent nets with attention modeling
  for ocr in the wild,'' in \emph{The IEEE Conference on Computer Vision and
  Pattern Recognition (CVPR)}, 2016.

\bibitem{Shi2016rar}
B.~Shi, X.~Wang, P.~Lyu, C.~Yao, and X.~Bai, ``Robust scene text recognition
  with automatic rectification,'' in \emph{The IEEE Conference on Computer
  Vision and Pattern Recognition (CVPR)}, 2016.

\bibitem{shi2019aster}
{Baoguang Shi}, {Mingkun Yang}, {Xinggang Wang}, {Pengyuan Lyu}, {Cong Yao},
  and {Xiang Bai}, ``Aster: An attentional scene text recognizer with flexible
  rectification,'' \emph{IEEE Transactions on Pattern Analysis and Machine
  Intelligence}, vol.~41, no.~9, pp. 2035--2048, 2019.

\bibitem{luo2019moran}
C.~Luo, L.~Jin, and Z.~Sun, ``A multi-object rectified attention network for
  scene text recognition,'' \emph{CoRR}, vol. abs/1901.03003, 2019.

\bibitem{Yang2019SCRN}
M.~Yang, Y.~Guan, M.~Liao, X.~He, K.~Bian, S.~Bai, C.~Yao, and X.~Bai,
  ``Symmetry-constrained rectification network for scene text recognition,'' in
  \emph{The IEEE International Conference on Computer Vision (ICCV)}, 2019.

\bibitem{Cheng2017FAN}
Z.~Cheng, F.~Bai, Y.~Xu, G.~Zheng, S.~Pu, and S.~Zhou, ``Focusing attention:
  Towards accurate text recognition in natural images,'' in \emph{The IEEE
  International Conference on Computer Vision (ICCV)}, 2017.

\bibitem{Cheng2018AON}
Z.~Cheng, Y.~Xu, F.~Bai, Y.~Niu, S.~Pu, and S.~Zhou, ``Aon: Towards
  arbitrarily-oriented text recognition,'' in \emph{The IEEE Conference on
  Computer Vision and Pattern Recognition (CVPR)}, 2018.

\bibitem{li2019sar}
H.~Li, P.~Wang, C.~Shen, and G.~Zhang, ``Show, attend and read: {A} simple and
  strong baseline for irregular text recognition,'' \emph{CoRR}, vol.
  abs/1811.00751, 2018.

\bibitem{VaswaniS2017transformer}
A.~Vaswani, N.~Shazeer, N.~Parmar, J.~Uszkoreit, L.~Jones, A.~N. Gomez, L.~u.
  Kaiser, and I.~Polosukhin, ``Attention is all you need,'' in \emph{Advances
  in Neural Information Processing Systems}, 2017, pp. 5998--6008.

\bibitem{wang2019src}
P.~Wang, L.~Yang, H.~Li, Y.~Deng, C.~Shen, and Y.~Zhang, ``A simple and robust
  convolutional-attention network for irregular text recognition,''
  \emph{CoRR}, vol. abs/1904.01375, 2019.

\bibitem{warps1989thin}
F.~L. B.~P. Warps, ``Thin-plate splines and the decompositions of
  deformations,'' \emph{IEEE Transactions on Pattern Analysis and Machine
  Intelligence}, vol.~11, no.~6, 1989.

\bibitem{simonyan2015vgg}
K.~Simonyan and A.~Zisserman, ``Very deep convolutional networks for
  large-scale image recognition.''\hskip 1em plus 0.5em minus 0.4em\relax
  Computational and Biological Learning Society, 2015, pp. 1--14.

\bibitem{He_2016_resnet}
H.~Kaiming, Z.~Xiangyu, R.~Shaoqing, and S.~Jian, ``Deep residual learning for
  image recognition,'' in \emph{The IEEE Conference on Computer Vision and
  Pattern Recognition (CVPR)}, 2016.

\bibitem{Szegedy_2015_inception}
C.~Szegedy, W.~Liu, Y.~Jia, P.~Sermanet, S.~Reed, D.~Anguelov, D.~Erhan,
  V.~Vanhoucke, and A.~Rabinovich, ``Going deeper with convolutions,'' in
  \emph{The IEEE Conference on Computer Vision and Pattern Recognition (CVPR)},
  2015.

\bibitem{Woo_2018_cbam}
S.~Woo, J.~Park, J.-Y. Lee, and I.~S. Kweon, ``Cbam: Convolutional block
  attention module,'' in \emph{The European Conference on Computer Vision
  (ECCV)}, 2018.

\bibitem{Liu2018CharNet}
W.~Liu, C.~Chen, and K.-Y.~K. Wong, ``Char-net: A character-aware neural
  network for distorted scene text recognition,'' in \emph{AAAI}, 2018.

\bibitem{liao2019scene}
M.~Liao, J.~Zhang, Z.~Wan, F.~Xie, J.~Liang, P.~Lyu, C.~Yao, and X.~Bai,
  ``Scene text recognition from two-dimensional perspective,'' in \emph{AAAI},
  vol.~33, 2019, pp. 8714--8721.

\bibitem{zhan2019esir}
F.~Zhan and S.~Lu, ``Esir: End-to-end scene text recognition via iterative
  image rectification,'' in \emph{IEEE Conference on Computer Vision and
  Pattern Recognition (CVPR)}, 2019.

\bibitem{gao12text}
L.~Yujia, G.~Hongchao, W.~Xi, H.~Jizhong, and L.~Ruixuan, ``Text recognition
  using local correlation,'' 2019.

\bibitem{Baek_2019_what}
J.~Baek, G.~Kim, J.~Lee, S.~Park, D.~Han, S.~Yun, S.~J. Oh, and H.~Lee, ``What
  is wrong with scene text recognition model comparisons? dataset and model
  analysis,'' in \emph{The IEEE International Conference on Computer Vision
  (ICCV)}, 2019.

\bibitem{liu2019safe}
W.~Liu, C.~Chen, e.~C.~V. Kwan-Yee K.~Wong", H.~Li, M.~Greg, and S.~Konrad,
  ``Safe: Scale aware feature encoder for scene text recognition,'' in
  \emph{ACCV}, 2019.

\bibitem{wan20192dctc}
Z.~Wan, F.~Xie, Y.~Liu, X.~Bai, and C.~Yao, ``2d-ctc for scene text
  recognition,'' \emph{CoRR}, 2019.

\bibitem{wang2020faclstm}
W.~Qingqing, H.~Ye, J.~Wenjing, H.~Xiangjian, M.~Blumenstein, L.~Shujing, and
  L.~Yue, ``Faclstm: Convlstm with focused attention for scene text
  recognition,'' \emph{Science China Information Sciences}, vol.~63, no.~2, p.
  120103, 2020.

\bibitem{ic03}
S.~LUCAS, A.~Panaretos, L.~Sosa, A.~Tang, S.~Wong, R.~Young, K.~Ashida,
  H.~Nagai, M.~Okamoto, H.~Yamamoto, H.~Miyao, Y.~Zu, W.~Ou, C.~Wolf, J.-M.
  Jolion, L.~Todoran, M.~Worring, and X.~Lin, ``{ICDAR 2003 Robust Reading
  Competitions: Entries, Results and Future Directions},'' \emph{{International
  Journal of Document Analysis and Recognition}}, vol.~7, pp. 105--122, 2005.

\bibitem{ic13}
D.~Karatzas, F.~Shafait, S.~Uchida, M.~Iwamura, L.~G. i~Bigorda, S.~R. Mestre,
  J.~Mas, D.~F. Mota, J.~A. Almazàn, and L.~P. de~las Heras, ``Icdar 2013
  robust reading competition,'' in \emph{12th International Conference on
  Document Analysis and Recognition (ICDAR)}, 2013, pp. 1484--1493.

\bibitem{ic15}
D.~{Karatzas}, L.~{Gomez-Bigorda}, A.~{Nicolaou}, S.~{Ghosh}, A.~{Bagdanov},
  M.~{Iwamura}, J.~{Matas}, L.~{Neumann}, V.~R. {Chandrasekhar}, S.~{Lu},
  F.~{Shafait}, S.~{Uchida}, and E.~{Valveny}, ``Icdar 2015 competition on
  robust reading,'' in \emph{2015 13th International Conference on Document
  Analysis and Recognition (ICDAR)}, Aug 2015, pp. 1156--1160.

\bibitem{IIIT5K}
A.~Mishra, K.~Alahari, and C.~Jawahar, ``{Scene Text Recognition using Higher
  Order Language Priors},'' in \emph{British Machine Vision Conference (BMVC)},
  2012.

\bibitem{svt}
K.~Wang, B.~Babenko, and S.~Belongie, ``End-to-end scene text recognition,'' in
  \emph{2011 International Conference on Computer Vision (ICCV)}, 2011, pp.
  1457--1464.

\bibitem{SVTP}
T.~Quy~Phan, P.~Shivakumara, S.~Tian, and C.~Lim~Tan, ``Recognizing text with
  perspective distortion in natural scenes,'' in \emph{The IEEE International
  Conference on Computer Vision (ICCV)}, 2013.

\bibitem{cute}
A.~Risnumawan, P.~Shivakumara, C.~S. Chan, and C.~L. Tan, ``A robust arbitrary
  text detection system for natural scene images,'' \emph{Expert Systems with
  Applications}, vol.~41, no.~18, pp. 8027 -- 8048, 2014.

\bibitem{zeiler2012adadelta}
M.~D. Zeiler, ``Adadelta: an adaptive learning rate method,'' \emph{arXiv
  preprint arXiv:1212.5701}, 2012.

\end{thebibliography}
%\begin{thebibliography}{1}
%\end{thebibliography}

\end{document}